\documentclass{article} % For LaTeX2e
\pdfoutput = 1
\usepackage{iclr2016_conference,times}
\usepackage{hyperref}
\usepackage{url}
\usepackage{amsmath}
\usepackage{subcaption}
\usepackage{graphicx}
\usepackage{makecell}

\title{Bounded Activation Functions: \\Towards Deep Binary Representations}
\title{Bounded Linear Functions: \\Towards Deep Binary Representations}
\title{Adjustable Bounded Rectifiers: \\Towards Deep Binary Representations}
\author{
Zhirong Wu, Dahua Lin, Xiaoou Tang \\
Department of Information Engineering\\
The Chinese University of Hong Kong\\
Shatin, Hong Kong \\
\texttt{\{zhirong,dhlin,xtang\}@ie.cuhk.edu.hk}
}

\begin{document}
%TODO
% give a nice title, how to name the activation function. --> talk to DH.
% rewrite experiments 3.
% think about any analysis of all binary networks --> code to exp 4.

\maketitle

\begin{abstract}
% motivation: I guess everybody should deduce what binary brings for the DNN.
Binary representation is desirable for its memory efficiency, computation speed and robustness.
%The ever-growing scale of deep neural networks often leads to high demand of computational resources. Among all techniques devised to tackle this problem, binarization, namely, turning continuous values into binary ones, is one of the most promising.
% introduce your topic and summarize the approach.
In this paper, we propose adjustable bounded rectifiers to learn binary representations for deep neural networks. 
%For The binary value only needs a single bit for storage and multiplicative computations are reduced to additions.
While hard constraining representations across layers to be binary makes training unreasonably difficult, we softly encourage activations to diverge from real values to binary by approximating step functions.
Our final representation is completely binary.
% brief summarization of contributions
% technical point
We test our approach on MNIST, CIFAR10, and ILSVRC2012 dataset, and systematically study the training dynamics of the binarization process. 
%may change this part based on later experimental results
Our approach can binarize the last layer representation without loss of performance and binarize all the layers with reasonably small degradations. 
The memory space that it saves may allow more sophisticated models to be deployed, thus compensating the loss.
To the best of our knowledge, this is the first work to report results on current deep network architectures using complete binary middle representations.
%We even manage to binarize the model parameters altogether with comparable results. 
%Given the learned representation, we find each binary neuron is attached with a semantic meaning by examining the input pattern of categories that consistently cause the neuron to fire or inhibit. 
% findings
Given the learned representations, we find that the firing or inhibition of a binary neuron is usually associated with a meaningful interpretation across different classes. This suggests that the semantic structure of a neural network may be manifested through a guided binarization process.
%We hope our model could bring great practical values on specialized hardwares as well as new insights about deep neural networks.
%Our approach is very easy to implement, and training parameters are also very easy to tune.
\end{abstract}

\section{Introduction}

% The good things about binary representation.
%Despite the computation power brought by the advent of GPUs, deep neural networks still suffer for wide deployment because of its large comsumption on storage and computations.
Despite the increased computational power resulted from the advances in high-performance computing, e.g. GPU and large-scale clusters, deployment of deep neural networks, especially under stringent resource constraints, remain a serious challenge because of their extraordinary demand on storage and computations. Binarizing the data representation along each layer would be a natural solution. In practice, most deep neural networks use single-precision (32-bit) floating point numbers for data representation. Hence, turning these values into binary ones would reduce the required storage by 32 times. Also, with real values replaced by binary values, the multiplication operations, which dominate the run-time in most cases, would reduce to bit-operations that are much more efficient.
%The ever-growing scale of deep neural networks often leads to high demand of computational resources. Among all techniques devised to tackle this problem, binarization, namely, turning continuous values into binary ones, is one of the most promising.
%The binary pattern of descriptors have long been desirable for its memory efficiency, computation speed and robustness. 
%Binarizing the representation along each layer would be a natural solution, 
%Specifically, it would drastically save the storage by 32 times from float to bit, and it replaces time-consuming multiplication operation with simple additions by constraining one of the multipliers to be 0 or 1. 
Overall, a binary representation would allow us to design much bigger and powerful networks using limited computation and storage resources. Moreover, as a way to understand the deep neural network, binary activations give a clear definition of neuron firing, which enables us to study the compositional architectures, the firing patterns, and various other properties in a principled way. 

%How it is generally a difficult problem, how we are going to approach it.
While we desire a complete binary representation, the optimization of an integer programming problem over millions of variables is intimidating. It is also somewhat unreasonable to constrain all the activations to be binary in the very beginning especially on lower layers. We instead take an alternative approach: the activations are initialized with real values, but are encouraged to diverge gradually during the training process. Hopefully, the network could reach to a solution where a binary representation is favored without the loss of performance. Concretely, our activation function is bounded between 0 and 1, and linear in between. The slope of the linear part is parameterized and can be learned from data. We encourage binary values by a weight growing constraint over these slope parameters (opposite to weight decay). Ideally, if the final activations are binary across all the data, the activation function works just like step functions. 
%Some remarks on the limitations
The limitation of the current approach is that the model consumes slightly more memory and computations during the training time. 

%Good properties of the bounded activation function
%Given the bounded rectifiers, one may worry the “vanishing gradients” problem will occur at the bounded regions. However, empirically, by initializing the slope parameters properly so that the input signal lies in the linear interval, the network will start to converge and will not trap in any bad local minimums using stochastic gradient descent. Since the activations of all layers are bounded to have almost the same magnitudes, the magnitude of parameter gradients across layers also tend to be more consistent. This allows us to use bigger learning rate for training. Furthermore, dropout are no longer necessary to combat overfitting. On the contrary, our weight growing constraint over the slope parameters can be seen as a new way to regularize the model since it encourages a robust representation. 

% technical contributions
We replace the ReLU nonlinearity with our bounded rectifiers in the current deep model architectures, and test the performance on MNIST, CIFAR10, and ImageNet standard benchmarks. We have the following findings.
First, our proposed activation function is more expressive than ReLU by observing no loss of performance if we don't force a complete binary representation. Since binary values are intrinsically more robust, we even find bounded activations as well as weight growing constraint could be considered as a new way to regularize the model.
Second, our approach can binarize the last layer representation without loss of performance on all the tasks, and suffer modestly when binarizing all layers in the network. We continue to study how each of the layers affects the performance when they are binarized. It turns out that binarization of some particular layers comprises most of the losses.
Third, when we enlarge the layers with more channels, we gain some obvious performance improvement. This indicates the configurations of some layers are too small and limit the performance for binary representations.
Last, since the binary represenation makes the output less sensitive to the model parameters, we even manage to binarize the model parameters altogether, which leads to a complete bit-wise deep model.

% insights findings
Binary neurons give a clear definition of fire and inhibition, which are a lot easier to be understood and interpreted than real values.
To understand these representations, we find that there exists some ``positive classes" and ``negative classes" where all the instances in the class consistently cause a binary neuron to turn on or off. 
In this sense, the function of this neuron is simply to separate  ``positive classes" and ``negative classes".  Some of the neurons are capable of representing high level concept like animals and articifials, while some other neurons are more elusive based only on class labels. 
To discover their functionality visually, we show the shared image pattern across these classes and assign a semantic description for the neuron. It turns out that the notion that these neurons have captured is still quite low level, such as repetitives, square shapes and rugged textures.
We believe these findings are truly intriguing. We hope that our binary model can help to study and understand the properties of deep neural networks, and potentially accelerate the deployment of DNN in computational limited environments.

%To analyze the divergence of each unit, we monitor its behavior through the training time. Given this information, we are able to identify some “dead neurons” which constantly turns on or off. Since these activations give no information to discriminate data, we suggest it’s a sign of overfitting and the network architecture should be shrinked. We also discover that, as a natural product of optimizing classification loss, binary representations in the middle layers tend to sparse. 

\section{Related Works}

%Binary representation as a feature in general?
Learning binary codes is an active research topic for machine learning in general. The code should be reasonably short, but contain as much information as possible, and allow fast computation during inference time. When handling tremendously large data, binary codes are essential for developing efficient search and matching algorithms. \citet{torralba2008small} first introduced the problem to the vision community. Techniques based on hashing such as  \citet{andoni2006near, weiss2009spectral} and quantization \citet{gong2011iterative} have been proposed. Encoding high-dimensional image data into short binary codes could be particularly useful for large-scale similarity search and indexing on the web. 
%For natural language processing, ... % Do we need something here?

%Binary representation in probabilistic models, DBN etc…
The binary representation for deep models is not a new topic. At the very beginning of neural network, inspired biologically, the heaviside step function has been used as the activation function. Training algorithms such as \citet{ toms1990training, barlett1992using} have been proposed. However, they never take full advantage of back-propagation algorithm, and can only work on toy examples with one hidden layer. For deep probabilistic models like RBM, DBN, DBM\citep{hinton2002training, hinton2006fast, salakhutdinov2009deep}, all of the internal representations are modeled as binary bernoulli variables. Despite a rather simple representation, these neural networks have shown that binary representations could still be very powerful for modeling complex data. However, for discriminative models especially convolutional neural networks that dramatically advance state-of-the-art in many areas, research that focuses on optimizing the binary codes is limited. 

%Observation: binary feature could be strong; fewer work chose to optimize it.
Previous work does make some positive observations by directly binarizing the features after the regular training is done. As the ReLU activation function has been widely used, the feature patterns can be understood as either activated or inhibited. In the paper \cite{agrawal2014analyzing}, they binarize the features as positives and zeros, and find that the performance drop is negligible. The experiment reveals that the local minimum of a binary representation for an individual layer exists. But it never discusses about how the binary representation could be further optimized and how binary representations for multiple layers coexist. Later on, \cite{lin2015deep, zhong2015deep} merely optimize the  representation of the last layer using a static sigmoid function, which is never truly binary. Our work enforces the representations to be truly binary and it can be applied throughout the whole network.

%Implementation methods, vanishing gradients, batchnorm, PReLU.
%Recent efforts on deep convolutional neural networks focus on exploring deeper and deeper network architectures \citep{simonyan2014very, szegedy2014going}. Initialization and optimization parameters play a crucial part in order for the network to converge. Techniques like batch normalization \citep{ioffe2015batch} and PReLU \citep{he2015delving} use first order and second order statistics to accelerate the convergence. In the paper \cite{ioffe2015batch}, they argue that a stable distribution of activation values is critical for training. In our case, the bounded functions confine the activations between 0 and 1, and somewhat have the effect of maintaining a stable distribution throughout training naturally. This allows us to use larger learning rate for training.

% deep model compression and binary weights
Our work is closely related to model compressions for deep neural networks as we are optimizing networks under constrained settings. \citet{han2015deep} reduces network parameters more than 30 times without performance loss by removing connection weights that below a certain threshold. As another way to compress networks, \cite{soudry2014expectation, courbariaux2015binaryconnect} propose deep networks with complete binary parameters. 
Unlike these work that focuses on compressing model parameters, we work on compressing the representations. This could be extremely useful for image applications where feature representations consume most of the storage.
We believe both lines of research hold the same promise to make the computation fast and storage consumption efficient. 

\section{Approach}
Our goal is to learn a binary representation for deep neural networks at some or all of the layers. Large-scale integer programming is extremely hard, especially for deep neural networks with millions of units. We avoid this approach, but choose to softly encourge the units from real values to binary values. We achieve this by designing an activation function that constrains the value between 0 and 1, and favors 0 and 1 as local minimums during training. The approach borrows the nice gradient property of ReLU and takes the full power of back-propagation for optimization. We now describe the activation function and show how we encourage the units to diverge.

\subsection{Adjustable bounded rectifiers}
\begin{figure}[t]
\centering
	\begin{subfigure}{0.24\textwidth}
			\centering
             \includegraphics[width=1\textwidth]{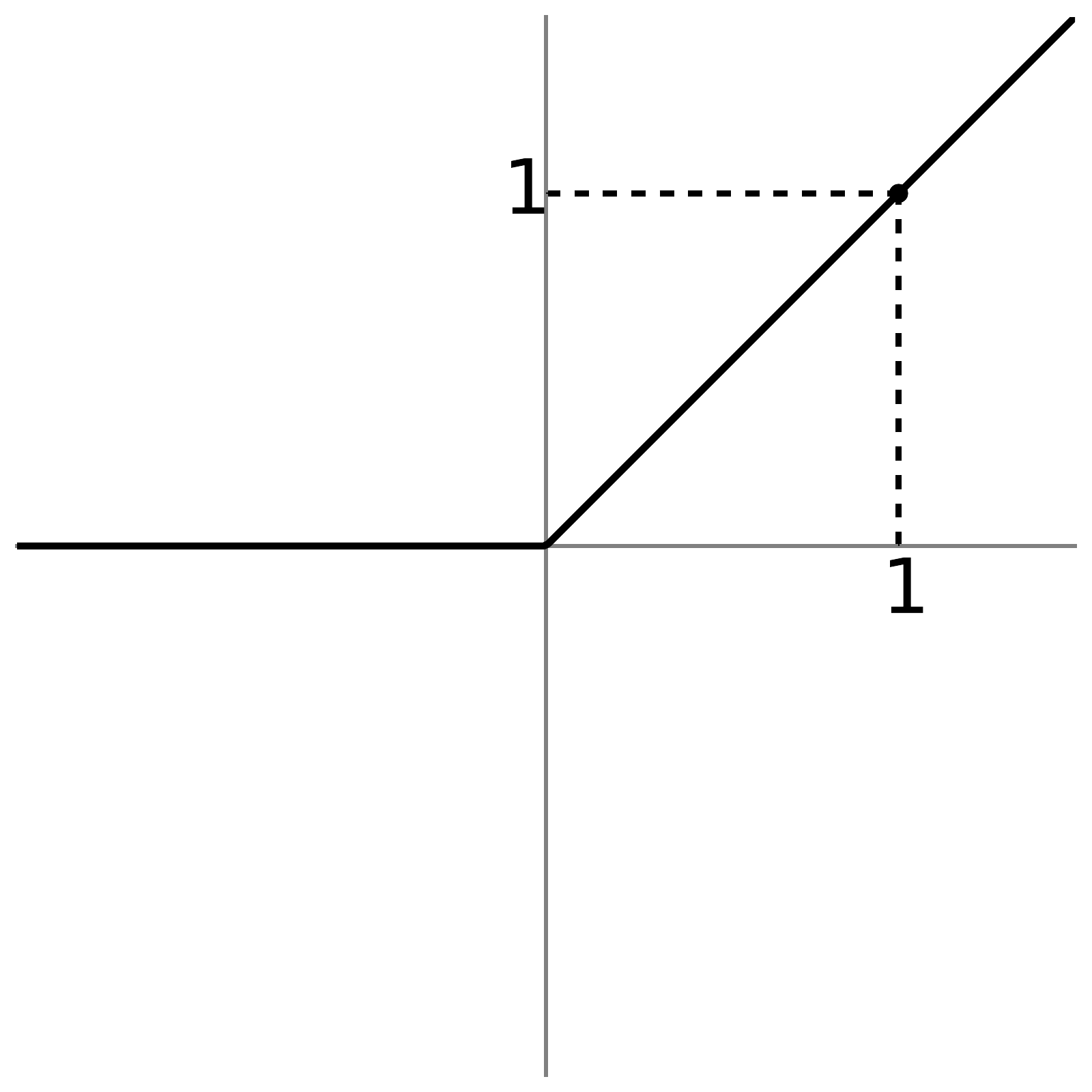}\\
              \caption{ReLU}
              \label{fig:relu}
     \end{subfigure}
     ~
     \begin{subfigure}{0.24\textwidth}
			\centering
             \includegraphics[width=1\textwidth]{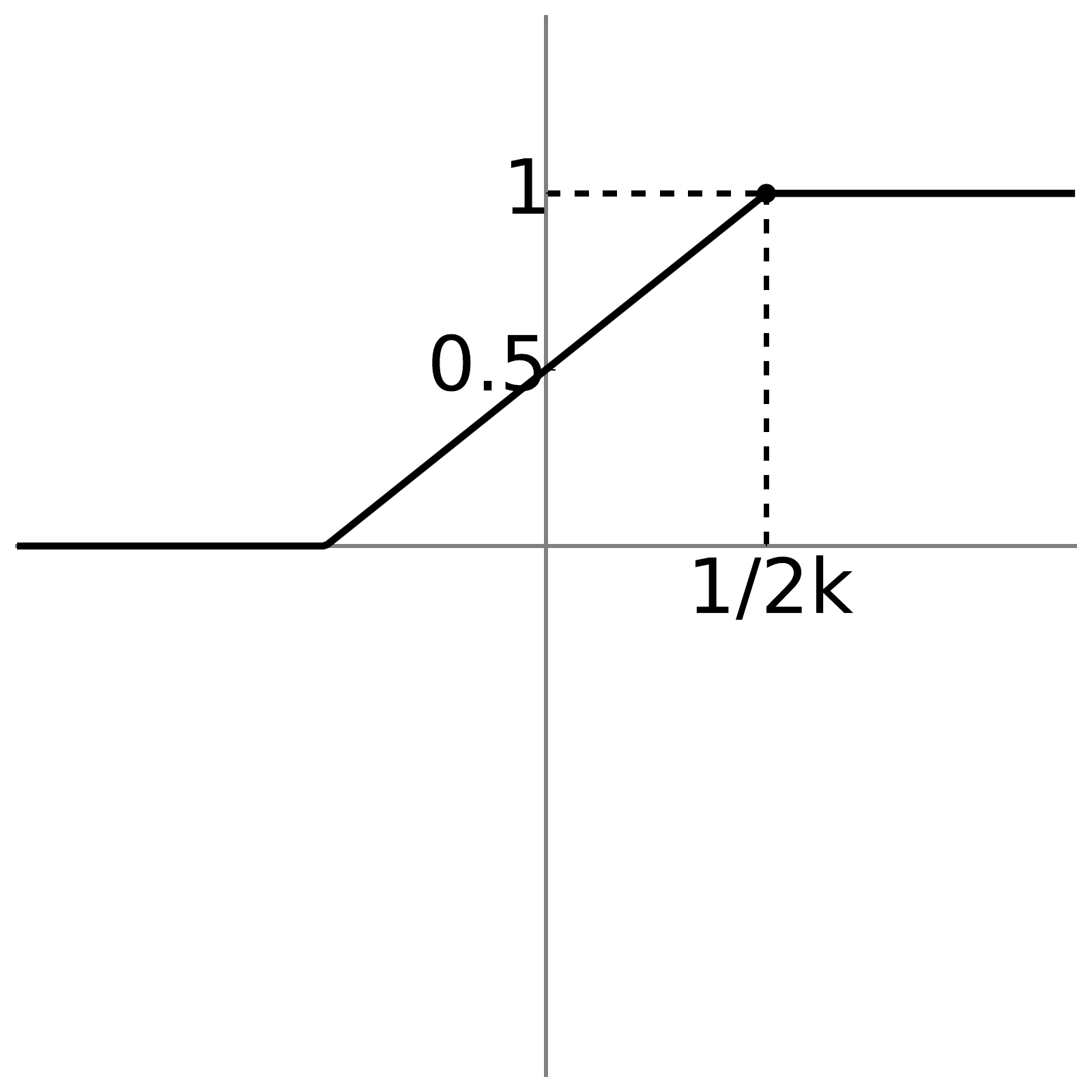}\\
              \caption{Bounded rectifiers}
              \label{fig:bounded}
     \end{subfigure}
     ~
	 \begin{subfigure}{0.45\textwidth}
			 \centering
             \includegraphics[width=1\textwidth]{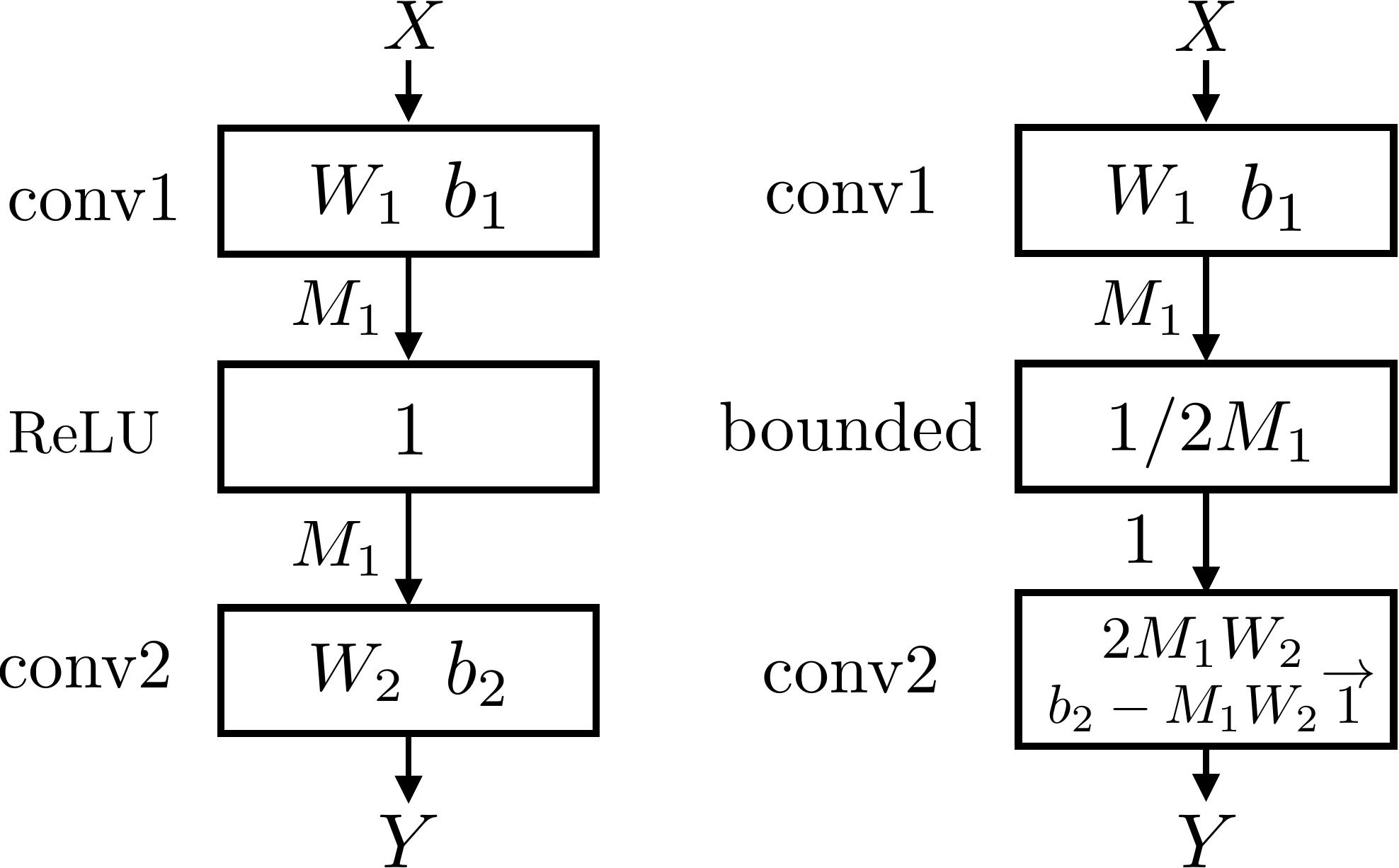}
              \caption{Functions expressed by a ReLU can always be cast into a module with bounded rectifiers.}
              \label{fig:relation}
	\end{subfigure}
\label{fig:activation_func}
\vspace{-2mm}
\caption{Adjustable bounded rectifiers and its relationship with ReLU.}
\vspace{-3mm}
\end{figure}

We design the activation function to be a simple linear function but clip the value less than 0 or bigger than 1. Formally, it can be defined as:
\begin{equation}
 f(y_i) = \text{min}\left(\text{max}\left(k_i y_i+0.5, 0\right),1\right)
\end{equation}
Where $y_i$ is the input of the activation function $f$ on the $i$th channel, $k_i$ is the coefficient controlling the slope of the linear interval. The subscript $i$ indicates that units in the same channel share the same slope parameters. When the slope $k_i$ is small, it allows the unit to perceive large input ranges; when the slope $k_i$ is large, it is only sensitive to small input ranges. When the activation $f(y_i)$ is saturated, it will have less chance of jumping back to the linear interval because the gradient in the saturated zone is 0. This is the intuition that we design an activation function which favors binary values.

Our activation function introduces a number of extra parameters, each slope coefficient for each feature channel. In fact, these parameters are redundant because they can be equivalently absorbed into the previous convolution or fully connected layers. Each set of parameters $W_i, b_i$ can be replaced by $k_iW_i, k_ib_i$,  and the activation function works with slope constant one.
%This could save unnecessary computations during inference time.
When the slope is large enough, our function degenerate close to the standard step function. The slope parameter can even be discarded and the activation function can be replaced by a step function during inference time.

One may worry the bounded rectifiers hurts the expressiveness of the neural network since it clips the values bigger than 1 compared with ReLU. However, the ``redundant" slope parameters actually make it even more powerful than ReLU. To see this, we can theoretically cast any function expressed by a ReLU network to a network with bounded activations. The idea is to set the slope so that the activations always lie in the linear interval. For a network with a convolution layer $W_{1}, b_{1}$, ReLU activation function and another convolution layer $W_{2}, b_{2}$. Suppose the maximum value of the first convolution layer is $M_{1} = \text{max}(|W_{1} X + b_{1}|)$. We can just set the slope in the new network to be $k_{1} = 1 / (2M_{1})$. In this way, the bounded rectifiers will never saturate with value bigger than 1. Since we shrink the magnitude of the network by $k_{1}$, we can compensate it by making the second convolution bigger $W_{2} \gets 2M_{1}W_{2} $ and do appropriate translations $b_{2} \gets b_2 - M_{1}W_{2} \overrightarrow{1}$. We can continue to do this for the rest of the network. Then our network with bounded rectifiers is equivalent to the ReLU network.

The adjustable bounded rectifiers are differentiable, and the slope parameters can be learned end-to-end together with other model parameters. The gradients of slope parameter $k_i$ is,
\begin{equation}
\frac{\partial E}{\partial k_i} = \sum_{y_i} \frac{\partial E}{\partial f{y_i}} \frac{\partial f(y_i)}{\partial k_i} = \sum_{y_i} \frac{\partial E}{\partial f(y_i)} y_i I(f(y_i) \leq 1) I(f(y_i) \geq 0)
\end{equation}
Where $E$ is the objective function of the neural network, $I$ is the identity function. The summation is applied on all the units $y_i$ that share the same slope parameter. With this derivation, we can train the whole neural network under the standard framework.

\subsection{Regularization As Weight Growth}
Deep neural networks usually have many local minimums. Although the units tend to trap in the saturated regions where no gradients is available, using the above activation function alone is not sufficient to have the desired binary representations. We have to aggressively encourage more units to diverge during the training process. One possible way is to interpret each activation as a probability, and add entropy loss for each individual unit. We find it generally works, but may be tricky to choose the loss weights across different layers. Another way is to let the bounded rectifiers get close to step functions. We can achieve this by encouraging the slope parameters to grow. This may sound strange at first, because parameters are prefered to be small to avoid overfitting. However, a bigger slope may lead to more binary values, which is robust and actually less prone to overfit. We even find that adding the slope growing constraint can lessen the necessity to use dropout. Concretely, we use the negative log as the loss function $-\text{log}(|k_i|)$. It has two appealing properties: First, the loss is unbounded towards the negative infinity. This would drive the slope $k_i$ to grow aggressively. Second, the derivative of the loss $1/k_i$  goes to zeros when the slope $k_i$ goes to infinity. This would make the optimization process stable. We have also experimented with other functions such as $1/|k_i|$ and $1/k_i^2$, but none of them works better than the log function.

\vspace{-1mm}
\subsection{Implementation Issues}

During the starting time of the training, the bounded rectifiers almost have the same behavior as the ReLU function. Therefore, any initialization methods that work for ReLU could be directly applied on the bounded rectifiers. In our experiments, we use the \textit{xavier} method \citep{glorot2010understanding} for all the layers (\citet{he2015delving} also works), and initialize the slope with value 1 (a gaussian with std value 1 also works). In the paper \cite{ioffe2015batch}, they manually normalize each layer to have zero mean and unit variance. This could be particular useful for networks that have dramatic training dynamics and have extremely different feature magnitudes. For example, the features of the VGG network \cite{simonyan2014very} are in the order of $10^2$ at the front while $10^{-1}$ at the end. However, our activation magnitude and training dynamics are much more stable because of the activation function. We just scale the input by $1/255$ to make the first layer consistent with the rest.

We divide the optimization protocol into two phases. In the first phase, we set the slope growing constraint to be relatively small. The constraint helps regularize the model, and the training procedure focuses on optimizing the performance, not on producing a binary representation. The first phase can reach the exact original performance with about 70 - 90\% binary representations (this may vary for different layers). In the second phase, we set the slope growing constraint to be a lot larger. This pushes the rest continuous values to binary while trying to maintain the objective performance. During the gradient update cycle, the growing constraint is added outside of the learning rate to aggressively binarize the representations. At the inference stage, we replace the bounded rectifiers with step functions.

\section{Experiments}
\label{exp}
In the following experiments, we work on MNIST, CIFAR10 \citep{krizhevsky2009learning}, and ImageNet ILSVRC2012 dataset \citep{deng2009imagenet}. The baseline methods are solvers and architectures
%\footnote{For the CIFAR10 architecture, there is no ReLU activation at the last layer originally. We add it as a baseline to better binarize that feature.} 
from the Caffe software package \citep{jia2014caffe}. We conduct the experiments on two primary settings: binarizing the features for the last layer, and binarizing the features for all layers. Both settings have great practical values, and our approach gives very promising results. We discuss binarizing model parameters altogether to derive a complete bit-wise deep model in the last section.

\subsection{Binarizing the last layer representation}
%1. Performance, baselines, binary evolution through time. Using Mnist, Cifar10.
%2. Robustness of Cifar10 last layer performance, trained longer time x4. (Maybe ImageNet).
\begin{figure}[t]
\centering
	\begin{subfigure}{0.45\textwidth}
			\centering
             \includegraphics[width=1\textwidth]{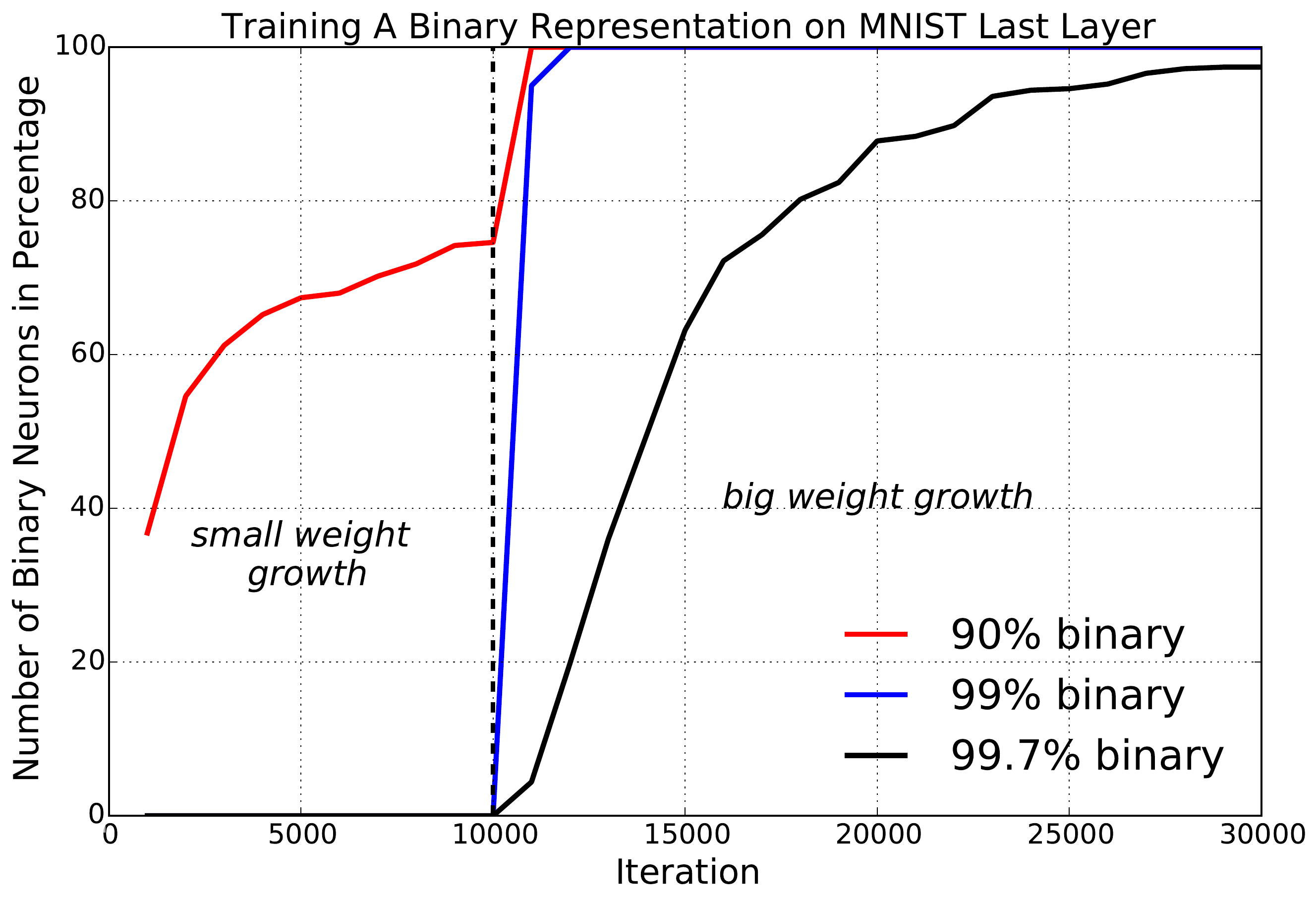}
              %\caption{Results on MNIST}
              %\label{fig:mnist_last}
     \end{subfigure}
     ~
     \begin{subfigure}{0.45\textwidth}
			\centering
             \includegraphics[width=1\textwidth]{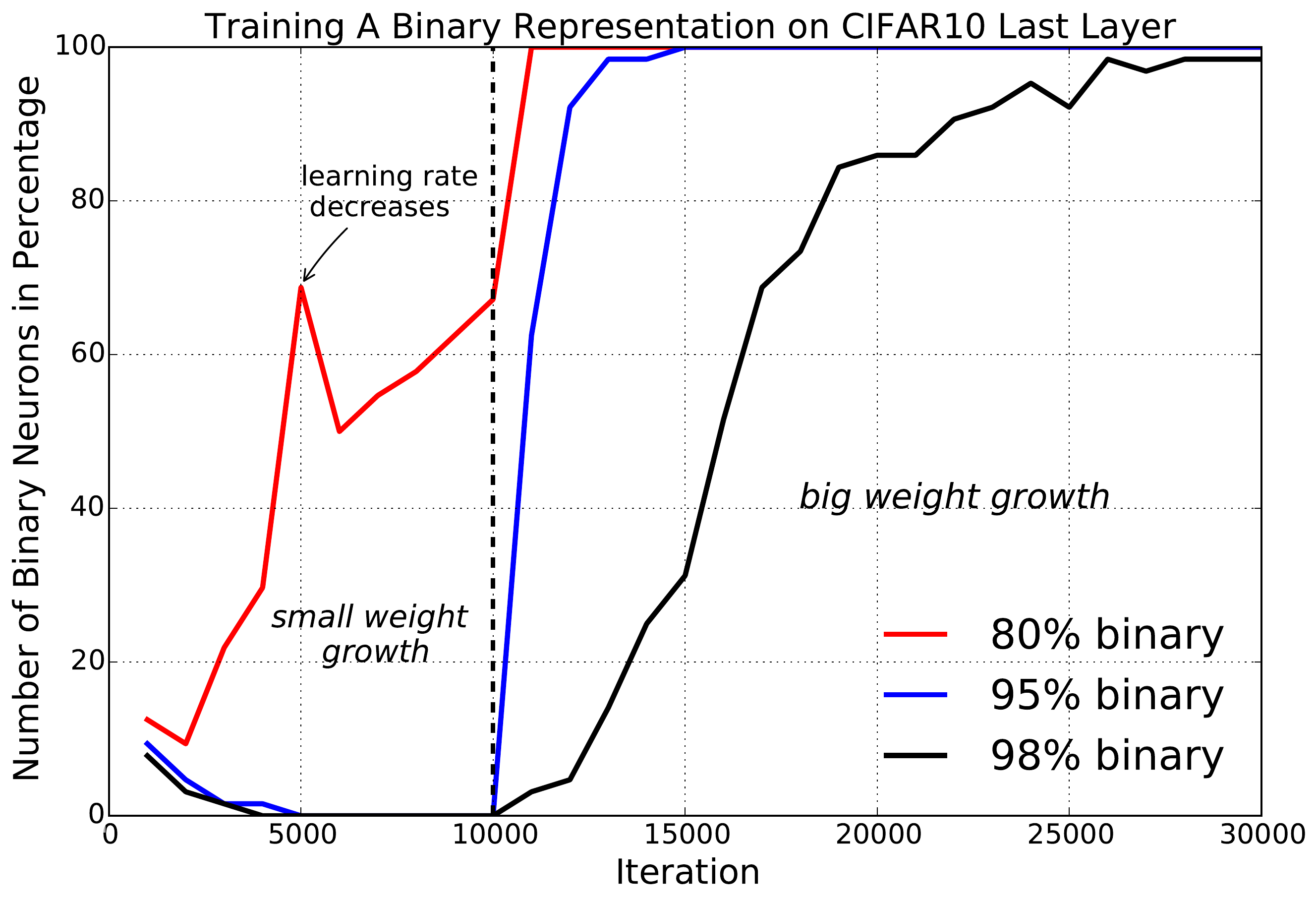}
              %\caption{Results on CIFAR10}
             % \label{fig:cifar10_last}
     \end{subfigure}
\vspace{-3mm}
\caption{Binarization process for the last layer along training time. Left: MNIST; Right: CIFAR10. For each curve in the figure, it is plotted as the percentage of total units which are at least $x\%$ (e.x. 90\%) of the time represented as binary values given all the testing data. We plot 3 curves under 3 different thresholds to better show the training dynamics. We find we are able to make 99.9\% units 99.9\% of the time binary if we train the network for unreasonably long time. However, it makes no difference for the final performance.}
\vspace{-2mm}
\label{fig:last_layer}
\end{figure}

\begin{table}[t]
\setlength{\tabcolsep}{3.6pt}
\centering
\caption{Classification performance for binarizing the last layer representations. ``Regular" means continuous feature values. ``Binary'' means directly binarizing the feature without retraining the softmax. ``Finetuned" means retraining the softmax.}
\vspace{-2mm}
\footnotesize
\begin{tabular}{c|c|c|c|c|c|c|c}
\Xhline{2\arrayrulewidth}
 & \multicolumn{3}{c|}{Baselines} & \multicolumn{4}{c}{Ours} \tabularnewline
\Xhline{2\arrayrulewidth}
 & Regular & Binary & Finetuned & 1st Phase & 1st Binary & 2nd Phase & 2nd Binary \tabularnewline
\Xhline{2\arrayrulewidth}
MNIST & 99.11& 95.38 & 98.37 & 99.21 & 99.12 & 99.22 & 99.22 \tabularnewline
\hline 
CIFAR10 &  75.40 & 38.04 & 62.39 & 75.60 & 74.10 & 75.96 &  75.76 \tabularnewline
\hline 
ImageNet & 56.66/79.92 & 53.67/77.86 & 54.06/78.15 & 56.22/79.54 & 54.69/78.13 & 56.23/79.53&  56.13/79.45 \tabularnewline
\Xhline{2\arrayrulewidth}
\end{tabular}
\label{table:last_layer}
\vspace{-3mm}
\end{table}

Binarizing the features for the last layer can be very useful for large scale image search and retrieval algorithms. 
Our experiment shows that features from the last layer of current CNNs can be safely binarized.
Take MNIST for an example, the baseline method obtains 99.11\% with standard ReLU activations. Directly binarizing the features without retraining the softmax suffers a lot. Retraining the classifier while fixing the binary features boosts the performance to 98.37\%. 
For our method, we monitor the level of binarization as well as the classification performance along the training time in Figure~\ref{fig:last_layer}.
Training the first phase using a small weight growth steadily binarizes more units. Direct binarizing the features without retraining the softmax gives a performance 99.12\%. Training the second phase using a larger weight growth maintains the accuracy 99.22\%, but substantially increases the level of binarization. Directly binarizing the features gives the same performance 99.22\%. Note there is no need to retrain the classifier for our method because the level of binarization is already very high. 
The corresponding experimental results on CIFAR10 and ImageNet are summarized in Table~\ref{table:last_layer}. From results on three datasets, we can always reach the performance using features as real values.
It is important to note that our method does not simply threshold ReLU activations, it learns a total different representation. For example, the Alexnet features on ImageNet are quite sparse, while ours is not sparse at all. Our feature has 58\% zero and 42\% one.
%We are not really comparing the absolute final performance, but emphasizing more on the difference

\begin{figure}[t]
\centering
	\begin{subfigure}{0.49\textwidth}
			\centering
             \includegraphics[width=1\textwidth]{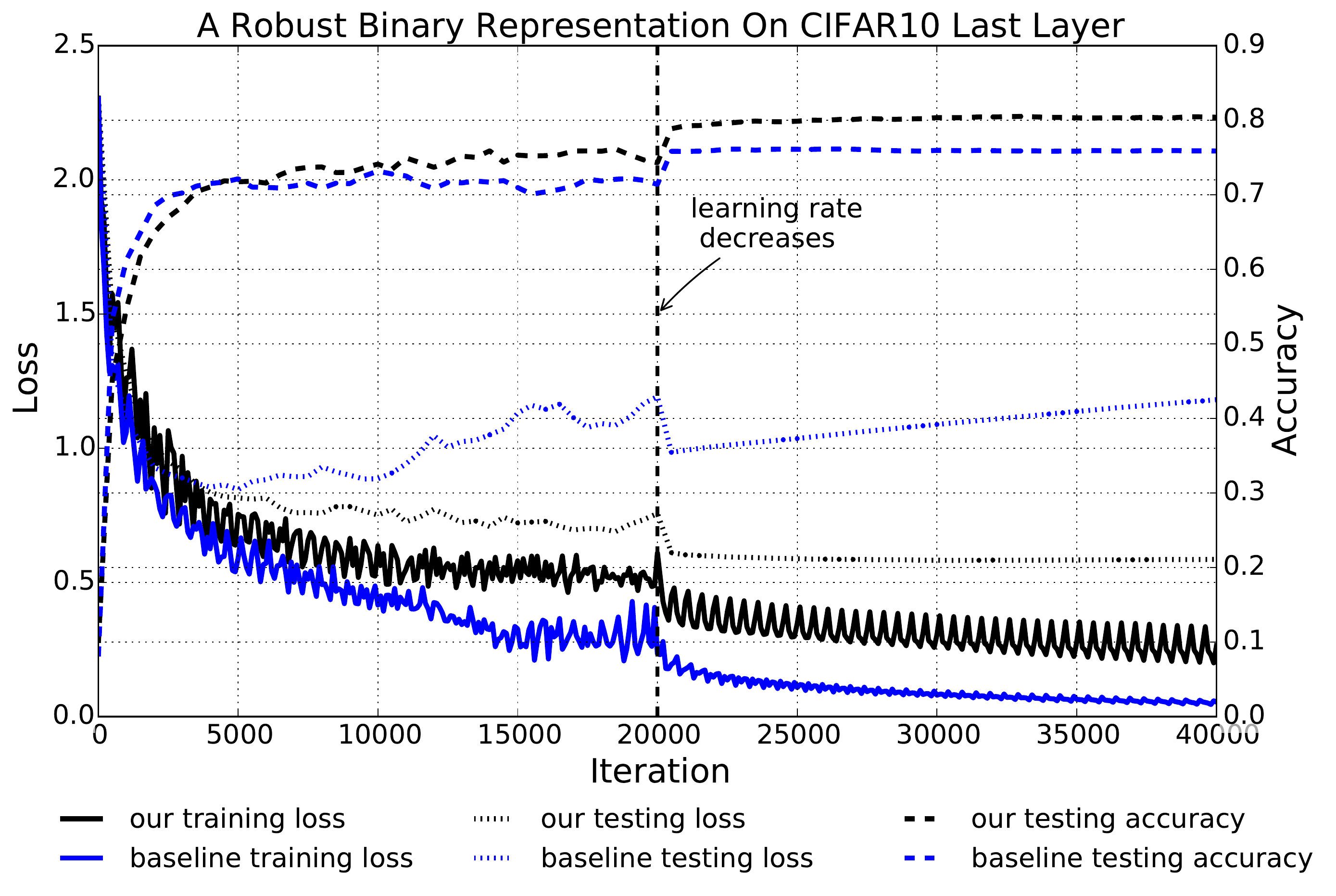}
              %\caption{Results on CIFAR10}
              %\label{fig:cifar10_robust}
     \end{subfigure}
     ~
     \begin{subfigure}{0.49\textwidth}
			\centering
             \includegraphics[width=1\textwidth]{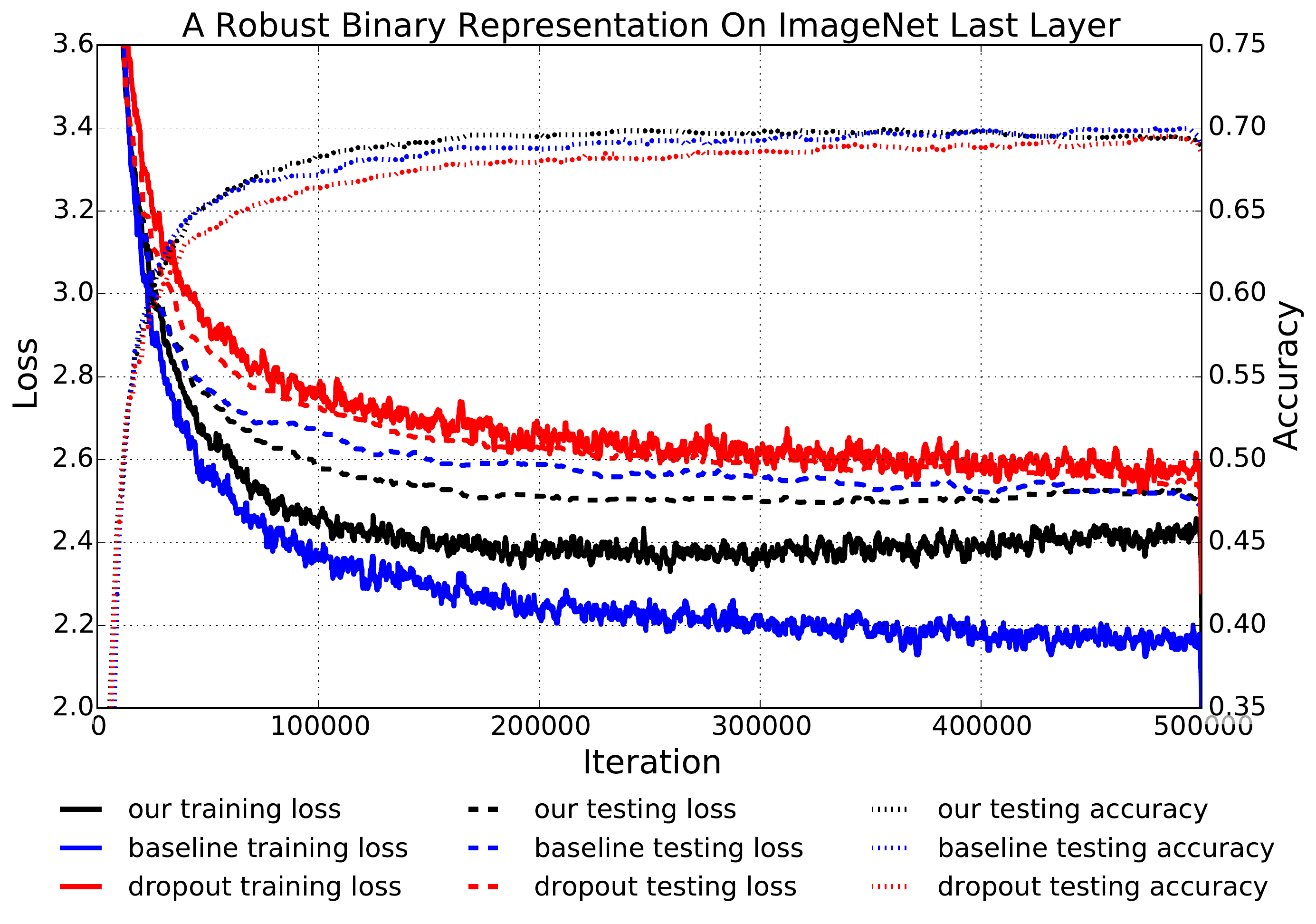}
             % \caption{Results on ImageNet}
              %\label{fig:imagenet_robust}
     \end{subfigure}
\vspace{-3mm}
\caption{Binary features are robust and less prone to overfit. Left: CIFAR10; Right: ImageNet. We monitor the training loss, the testing loss, the testing accuracy for longer training iterations. The gap of our training loss and testing loss is a lot smaller than ReLUs . Please note that our representation in this experiment is not completely binary.}
\label{fig:robust}
\vspace{-4mm}
\end{figure}

\textbf{Feature robustness.}
Binary features are inherently robust and less prone to overfit. To prove this, we train the deep models for longer iterations using a small weight growth constraint. In Figure~\ref{fig:robust}, the baseline model with ReLU activations is severely affected by overfitting, and the training loss drops a lot faster than the testing loss.
%We then replace ReLU with bounded rectifiers for the last layer while keeping other training paratemers fixed. In Figure~\ref{fig:robust}, we can see that a binary representation is robust to overfitting.
For our bounded rectifiers, on CIFAR10, longer training continues to give us better performance. We reach a performance of 80.48\%, which is significantly better than 75.40\%, which is the best performance that current method can achieve with the same architecture.
On ImageNet, our method only uses dropout on fc6 layer but not fc7 layer. One baseline setting is consistent with ours, and the other uses full dropout training.
%the baseline model only has dropout on the previous layers but not the last layer. We plot another curve for full dropout training.
This experiment shows that even when a binary representation is not desired, bounded rectifiers as well as weight growth constraint should be considered as an effective way for regularization.

%\vspace{-3mm}
\subsection{Binarizing the whole network}
%\vspace{-4mm}
%3. Cifar10 all layers binary. Change Network architecture. Make it bigger so that complete binarization still have the state-of-the-art performance.

\begin{figure}[t]
\centering
	\begin{subfigure}{0.50\textwidth}
			\centering
             \includegraphics[width=1\textwidth]{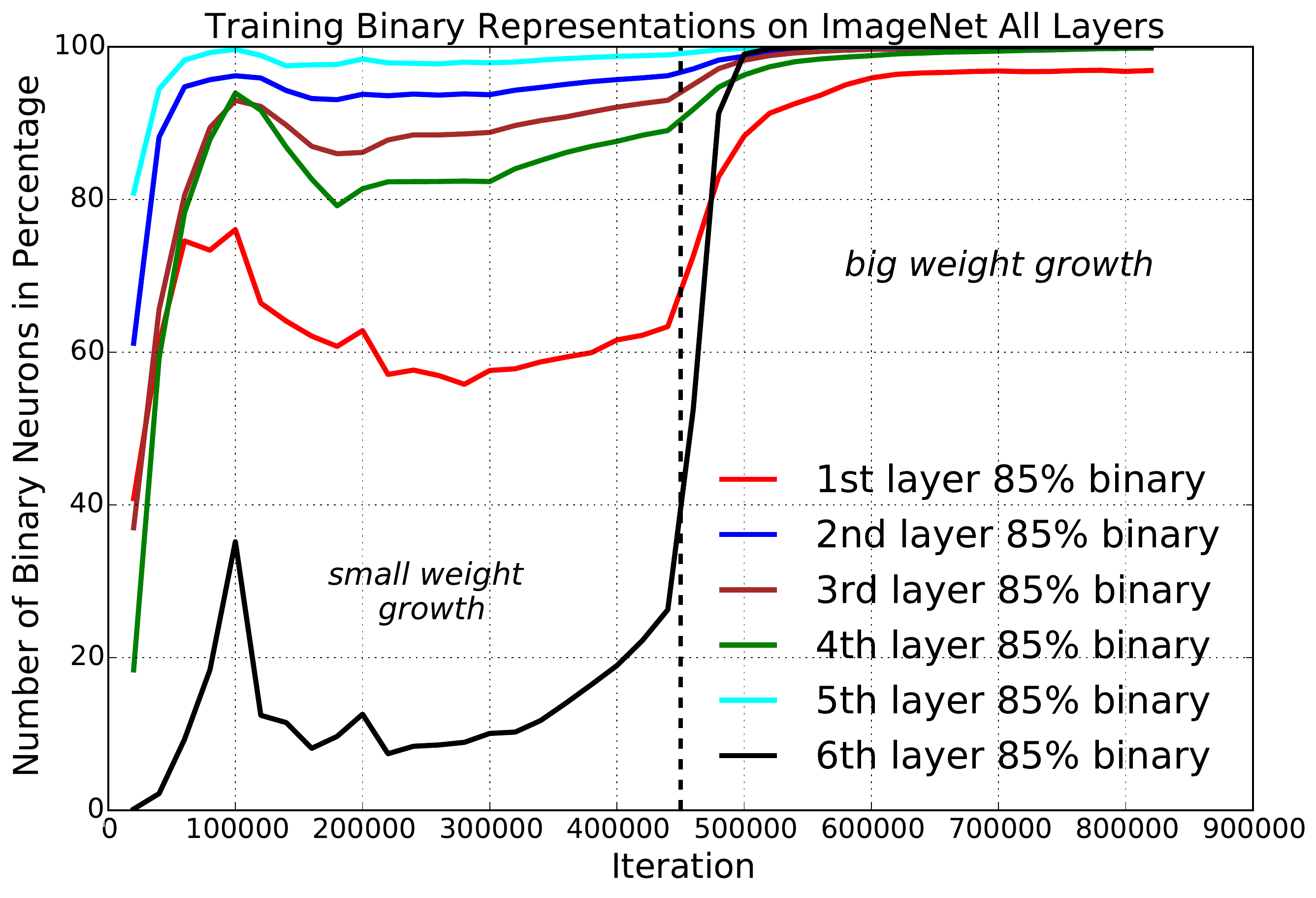}
             % \caption{Results on CIFAR10}
             % \label{fig:cifar10_all}
     \end{subfigure}
     ~
     \begin{subfigure}{0.38\textwidth}
			\centering
             \includegraphics[width=1\textwidth]{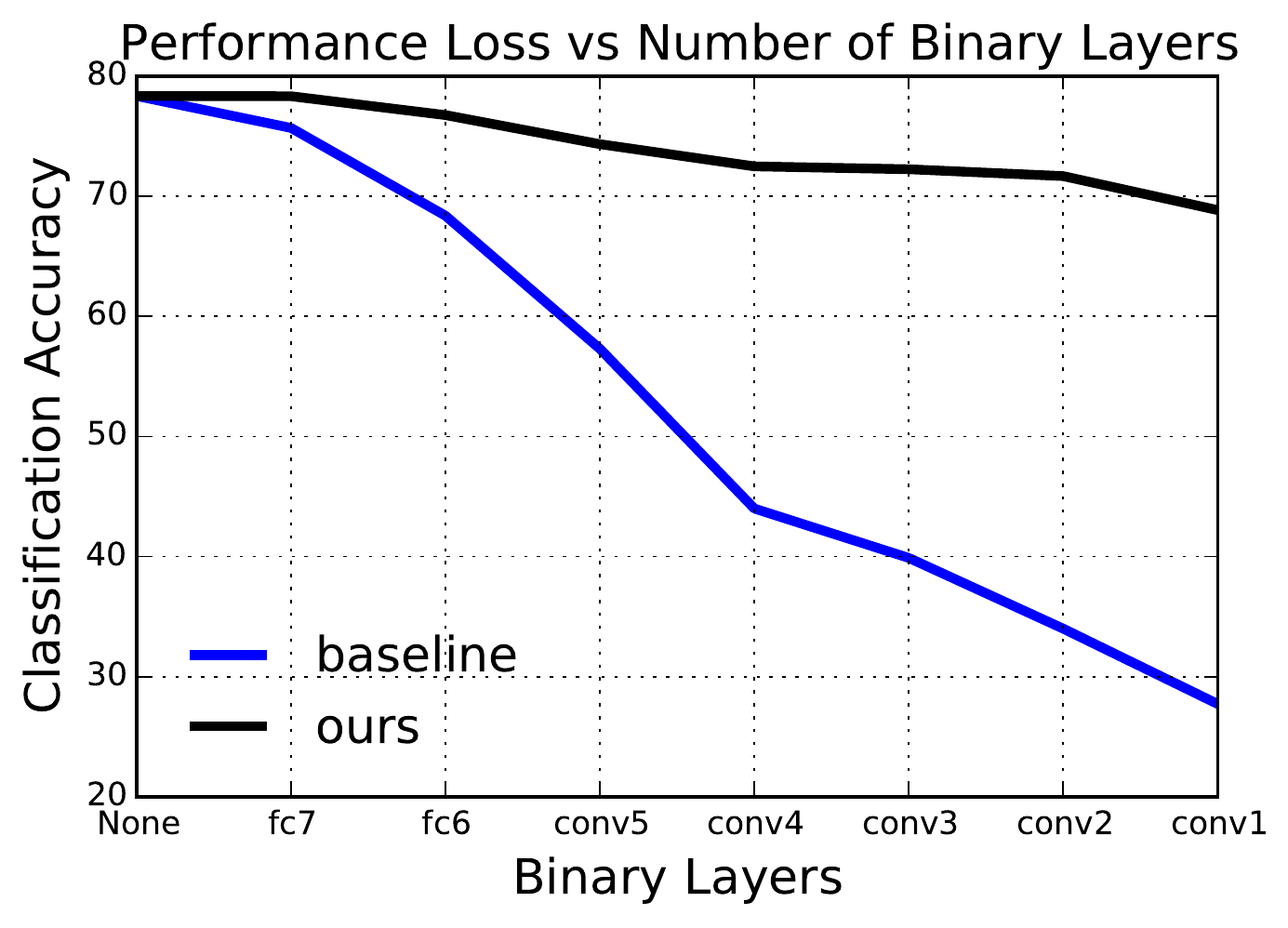}
             % \caption{Results on ImageNet}
             % \label{fig:imagenet_all}
     \end{subfigure}
\vspace{-3mm}
\caption{Left: Binarization process for all layers vs. training iterations on ImageNet. The curves are plotted in the same way as Figure~\ref{fig:last_layer}. We only plot one curve for each individual layer for simplicity. Naturally, the last layer is binarized slower than previous layers to allow gradient propagations. Right: Classification performance as the network has more binary layers using AlexNet on ImageNet. Conv2, conv3, fc7 layers exhibit no loss of performance when they are binarized. }
\label{fig:all_layers}
\vspace{-3mm}
\end{figure}

Binarizing the representations across all the layers could potentially save memory usage to a great extent and speed up computations. It also allows us to explore even deeper and wider network architectures. Apparently, forcing all the layers to be binary is an extremely difficult task. If the layer at the front is binarized while losing too much information, the following layers will have no chance to recover. Also, if the layer at the end is binarized a lot faster than previous layers, the gradient signal will be too weak to be propagated. To our surprise, in Figure~\ref{fig:all_layers}, our network can automatically maintain a steady level of binarization across all the layers while conservatively encouraging binarization.

We conduct experiments on CIFAR10 and ImageNet. For the baseline method\footnote{The baseline here is a little bit lower than original Alexnet because this baseline is trained with \textit{xavier} initializations. We want to cancel out the effects of initializations since the original Alexnet is sophisticatedly tuned.}, we take the well-trained ReLU model and it suffers a lot if we binarize all the layers simultaneously. Optimally, we can add a binary representation layer each time from the front to the back, and finetune all the layers behind as real values. Then we repeat this process until all the layers are binarized. The performance increases to 45.48\% and 27.73\%, respectively. For our approach, we replace ReLU activations with our bounded activations for all the layers and train the deep model from scratch. After two phases of training, we obtain an accuracy of 72.14\% and 68.39\%  by directly replacing each bounded rectifiers with a step function. 
%The performance can be further finetuned by observing the fact that the representation is still not completely binarized and the model is not fitted to a complete binary representation.
We can also follow the layer-wise finetuning as for the baseline method.  In our experiment, to make things simple, we just finetune the final softmax classifier. Our final result is 73.08\%, 68.85\%. Details are included in Table~\ref{table:all_layers}. 

\begin{table}[t]
\setlength{\tabcolsep}{1.6pt}
\centering
\footnotesize
\caption{Classification performance for binarizing the whole network representations. Key words are consistent with Table~\ref{table:last_layer}, except the baseline is finetuned layer-wise. Performance of the binary representation increases a lot when the architecture is made twice larger.}
\vspace{-2mm}
\begin{tabular}{c|c|c|c|c|c|c|c|c}
\Xhline{2\arrayrulewidth}
 & \multicolumn{3}{c|}{Baselines} & \multicolumn{5}{c}{Ours} \tabularnewline
\Xhline{2\arrayrulewidth}
 & Regular & Binary & Finetuned & 1st Phase & 1st Binary & 2nd Phase & 2nd Binary & Finetuned\tabularnewline
 \Xhline{2\arrayrulewidth}
CIFAR10 & 75.40  & 18.15 & 45.48 & 75.35 & 61.87 & 73.77 & 72.14 & 73.08 \tabularnewline
\hline
2x model & ---  & --- & --- & 78.05  & 69.05  & 77.25  & 75.71 & 76.22 \tabularnewline
%\hline
%3x model & ---  & --- & ---  &  &  &  &  &  \tabularnewline
 \Xhline{2\arrayrulewidth}
ImageNet & 55.10/78.35 & 0.1/0.5 & 12.44/27.73 & 53.81/77.74 & 30.24/54.13 & 46.64/71.63 & 43.33/68.39 & 43.70/68.85\tabularnewline
\hline
2x model & ---  & --- & --- & 57.87/80.93 & 39.81/63.67 & 54.07/77.73 & 49.51/73.61 & 49.85/74.23 \tabularnewline
%\hline
%3x model & ---  & --- & --- & & & & \tabularnewline
\Xhline{2\arrayrulewidth}
\end{tabular}
\label{table:all_layers}
\vspace{-4mm}
\end{table}

\textbf{Larger network architectures.}
We lose about 2.4\%, 9.5\% of accuracy by using the same architecture as networks with real values. To find out which layers our approach suffers most, we train a set of models that gradually have more binay layers, e.g. binarizing only the last layer, and add the second last layer, etc. In figure \ref{fig:all_layers}, on AlexNet, we can see that conv2, conv3, and fc7 layers do not hurt the performance while other layers may have some noticeable losses. Maybe it's the capacity of the network architecture that limits the performance for binary representations. Can we reach to the same performance or even better performance by using bigger networks? To test the possibility, we double the number of feature channels for all the layers, and redo the experiment under the same settings. Finally, we reach a performance of 76.22\% and 74.23\% for the twice larger networks.  
This obviously confirms the architecture should be made larger for binary values. Also, it would be very interesting to investigate the minimum binary architecture in order to match the original performance. For example, maybe only conv1 and conv5 layer needs to be made larger, and all the other layers could be remained the same.  Since it needs an exhaustive search over a large amount of models, we leave it for the  future work. 
%Also, our approach allows a hybrid representation of continuous values and binary values. One can carefully tune the network by a trade off between performance and binary representations. 

%The above experiments show activation units can be steadily binarized over training time. In this section, we investigate which units are binarized first and which are binarized last for each feature map. Units that are binarized last tend to be the most difficult, and are more likely to contain the discriminative information. Concretely, we visualize the activation response map of each layer across entire training time. 

\subsection{Activation patterns of binary feature units}

\begin{figure}[t]
\centering
	\begin{subfigure}{0.98\textwidth}
			\centering
             \includegraphics[width=0.8\textwidth]{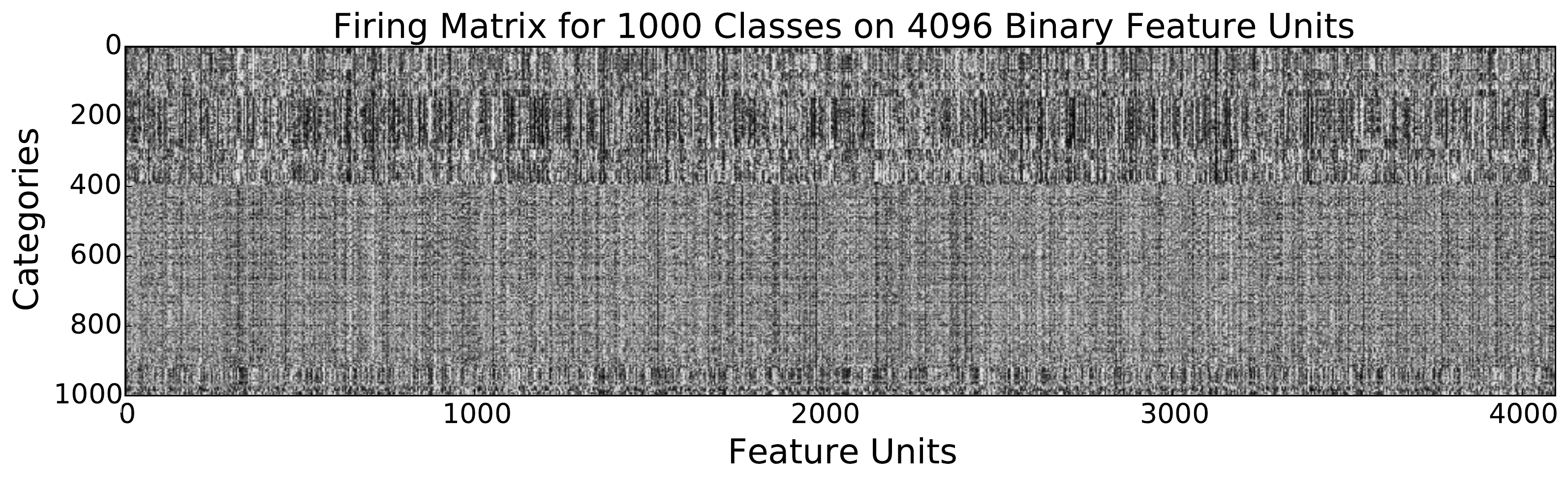}
              \label{fig:fire_matrix}
     \end{subfigure}
     \begin{subfigure}{1.0\textwidth}
			\centering
             \includegraphics[width=0.245\textwidth]{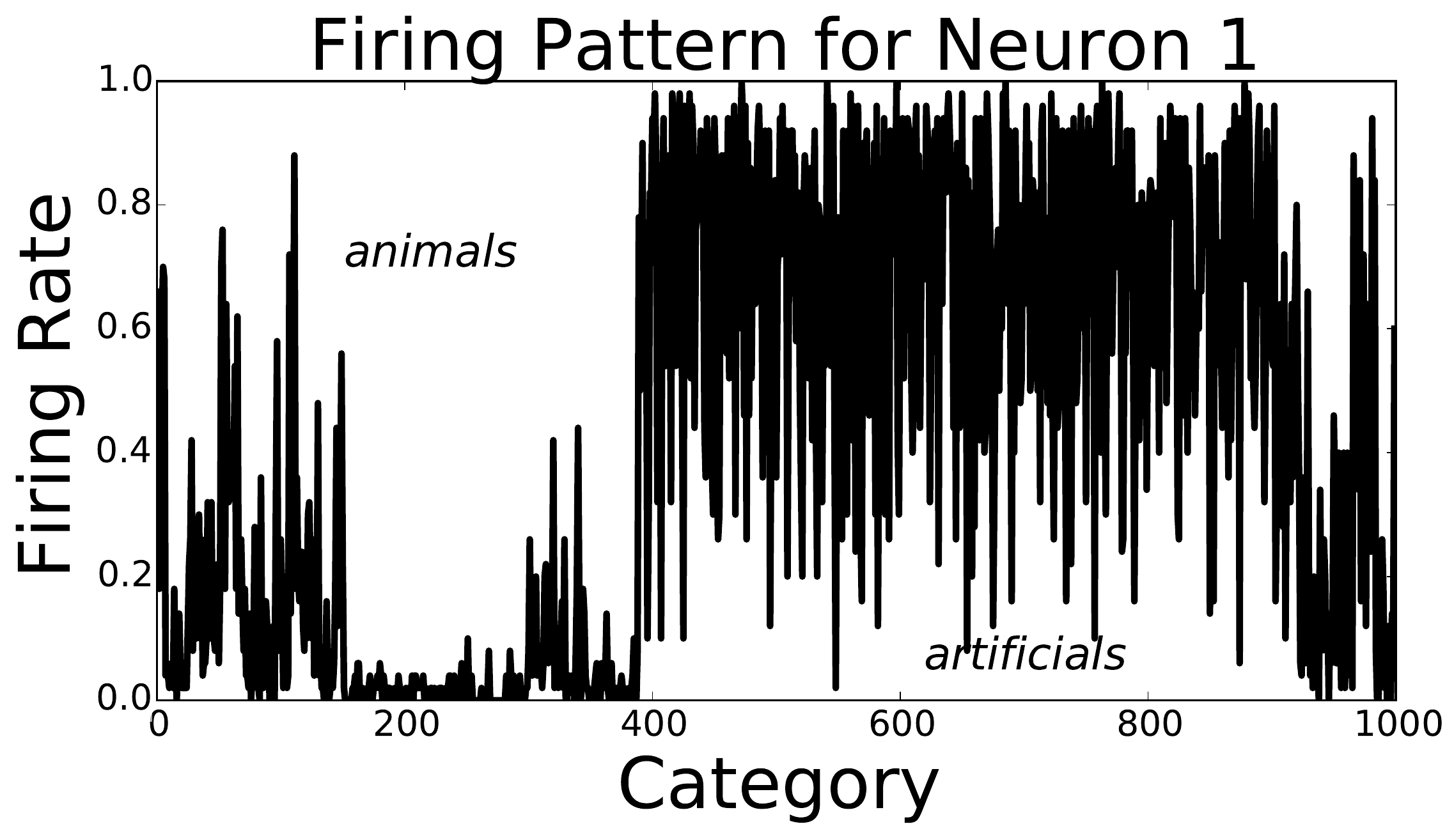}
             \includegraphics[width=0.245\textwidth]{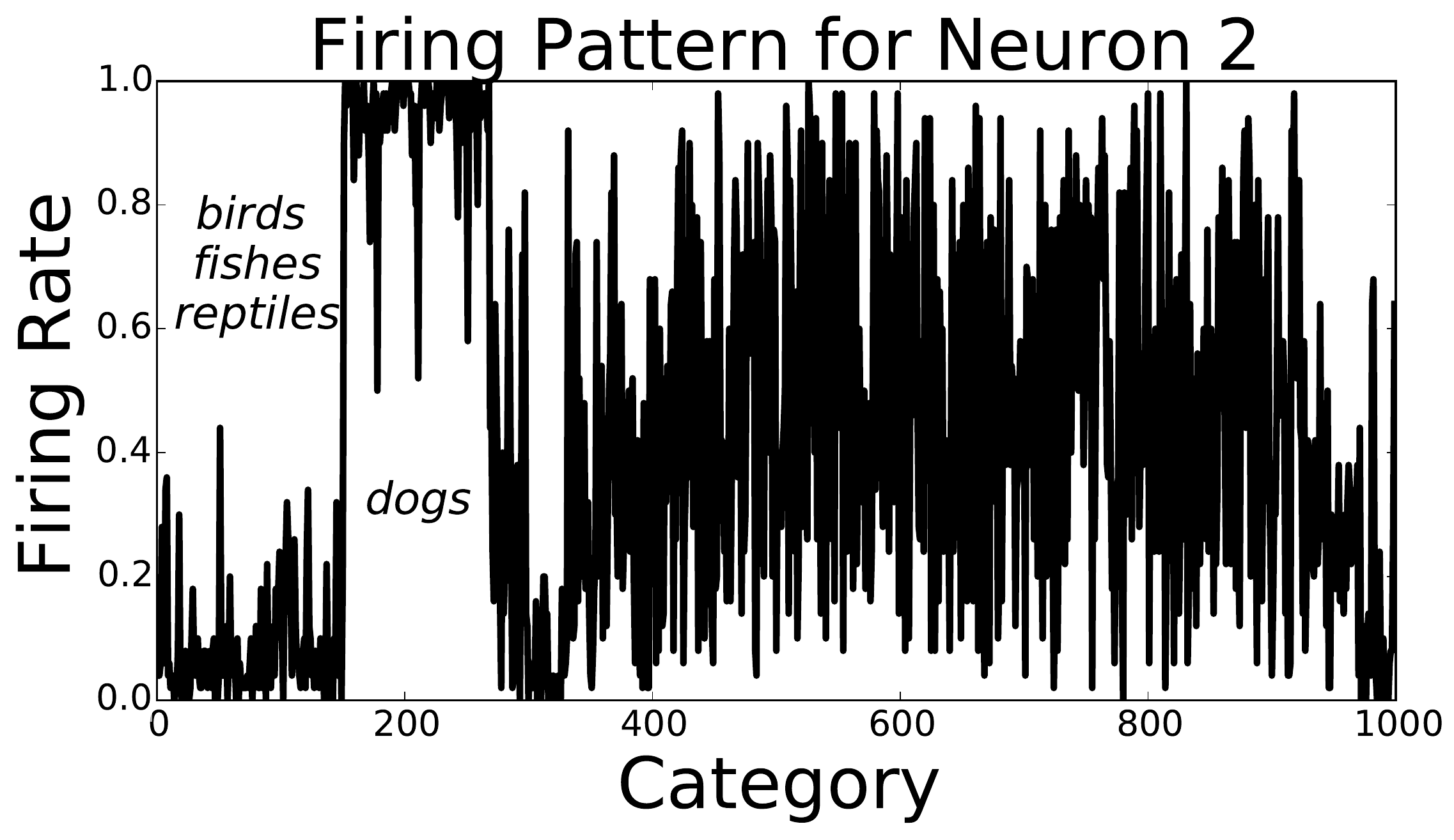}
             \includegraphics[width=0.245\textwidth]{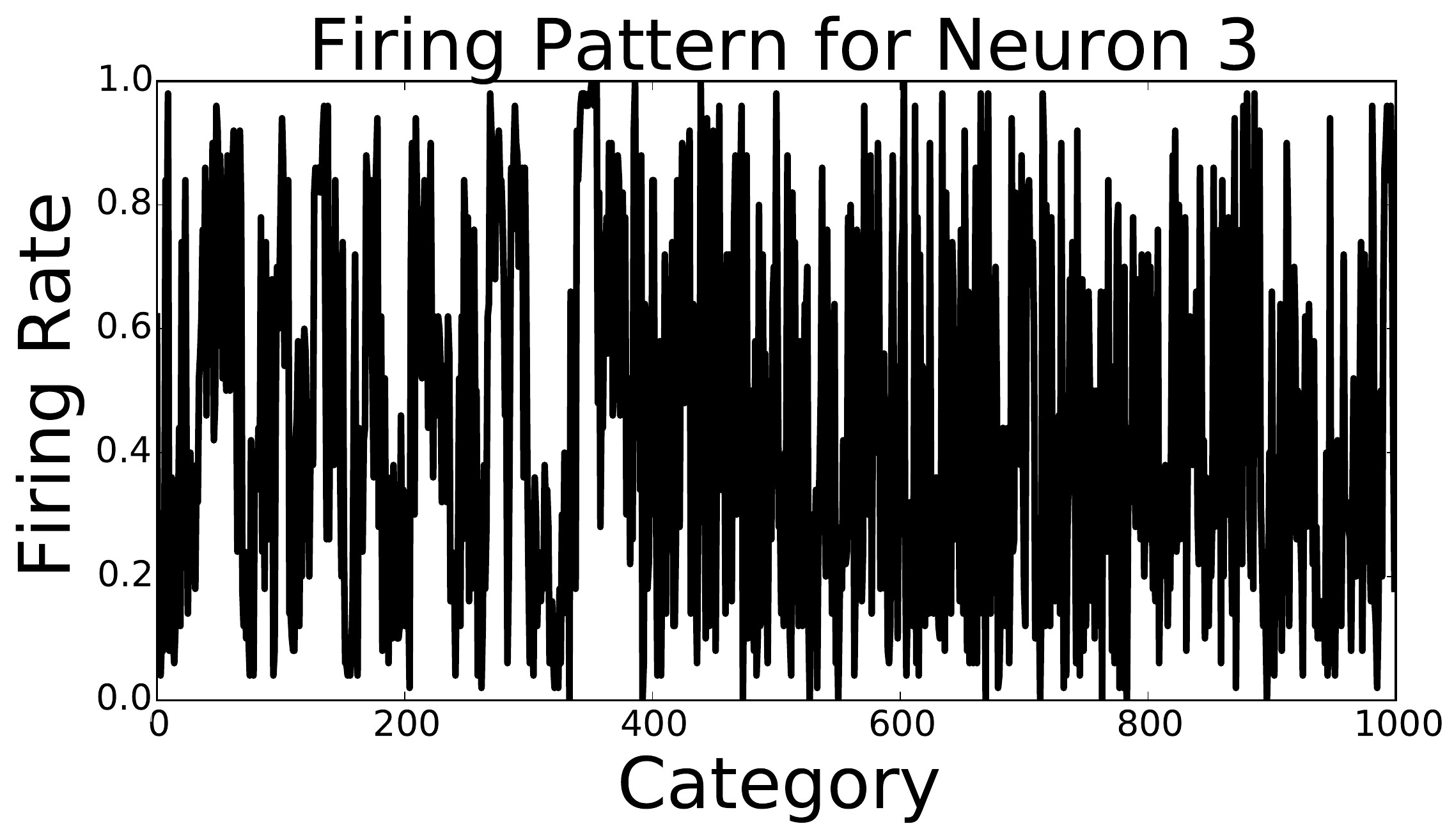}
             \includegraphics[width=0.245\textwidth]{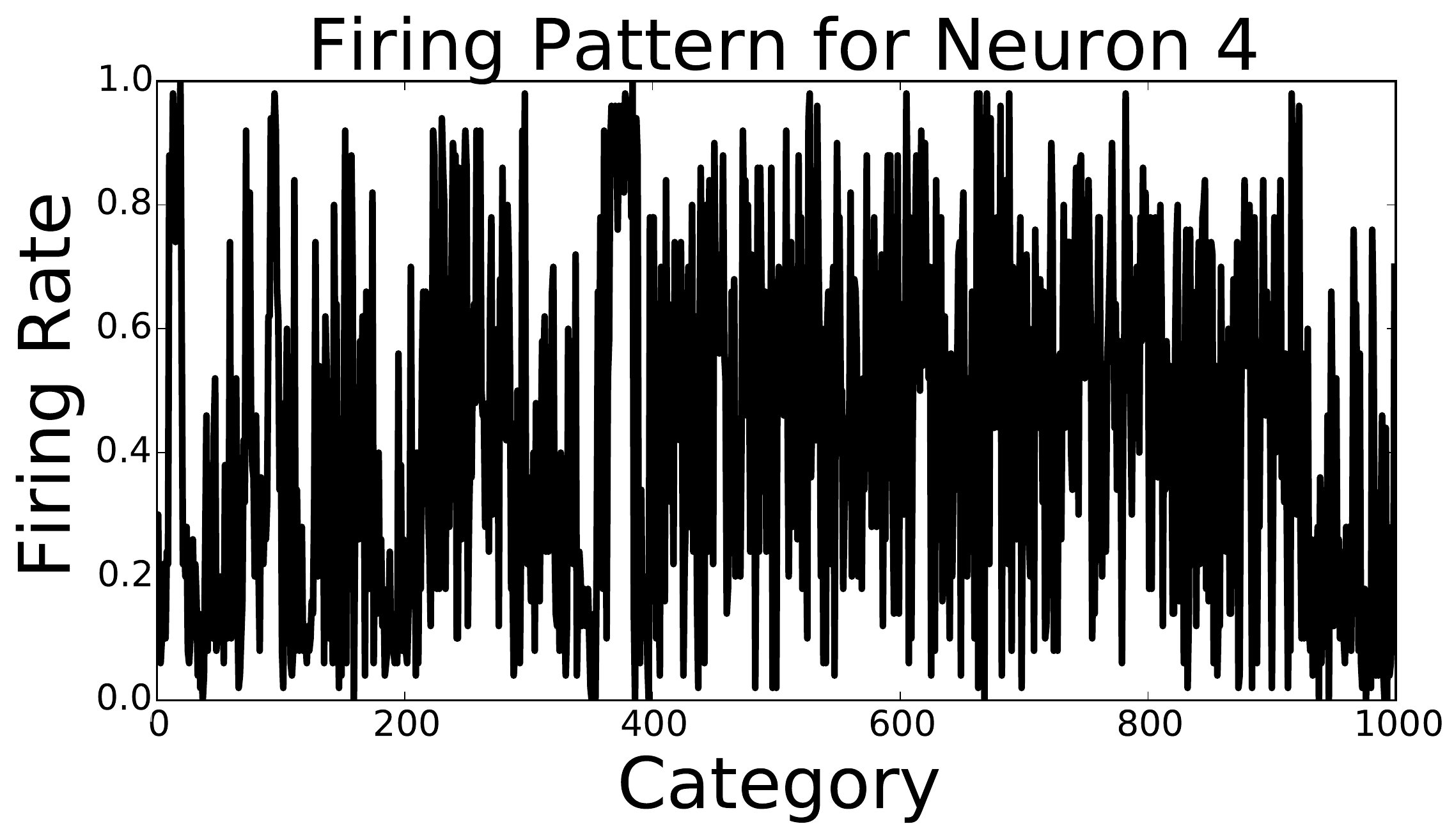}
              \label{fig:neuron_semantics}
     \end{subfigure}
\vspace{-4mm}
\caption{Top: Firing matrix for each class on each feature unit. Each element indicates how often a class fires for a particular binary unit. Bottom: A column of the firing matrix reveals an activation distribution for one unit over all categories. Some units are clearly related to semantics. e.g. Unit 1 is used to distinguish animals and artificials, and unit 2 is used to distinguish dogs and birds/fishes/reptiles. Some units (3 and 4) are not obviously semantically related.}
\label{fig:fire_pattern}
\vspace{-6mm}
\end{figure}

Binary representations give a clear definition of fire and non-fire. Can we develop an interpretation of the binary value zero and one? 
To start with, we work on ImageNet dataset and calculate the average responses of the fc7 units for all 1000 classes. This gives us a $1000\times 4096$ matrix (see Figure \ref{fig:fire_pattern}) where each element indicates how often each class fires for a specific neuron. We find that for some neurons, there exists some ``positive classes" where \textit{all} the instances in the class consistently fire for this neuron, and some ``negative classes" where \textit{all} the instances in the class consistently inhibit for this neuron. Such neurons carry strong discriminative information for multi-class classification. In fact, given only one such particular neuron, we can easily build a classifier to separate ``positive classes" and ``negative classes" by first adding a bias 0.5 and choosing a weight of 1 for positive classes, -1 for negative classes, and 0 for the remaining ambiguous classes. 

\begin{figure}[h]
\centering
\includegraphics[width=0.90\textwidth]{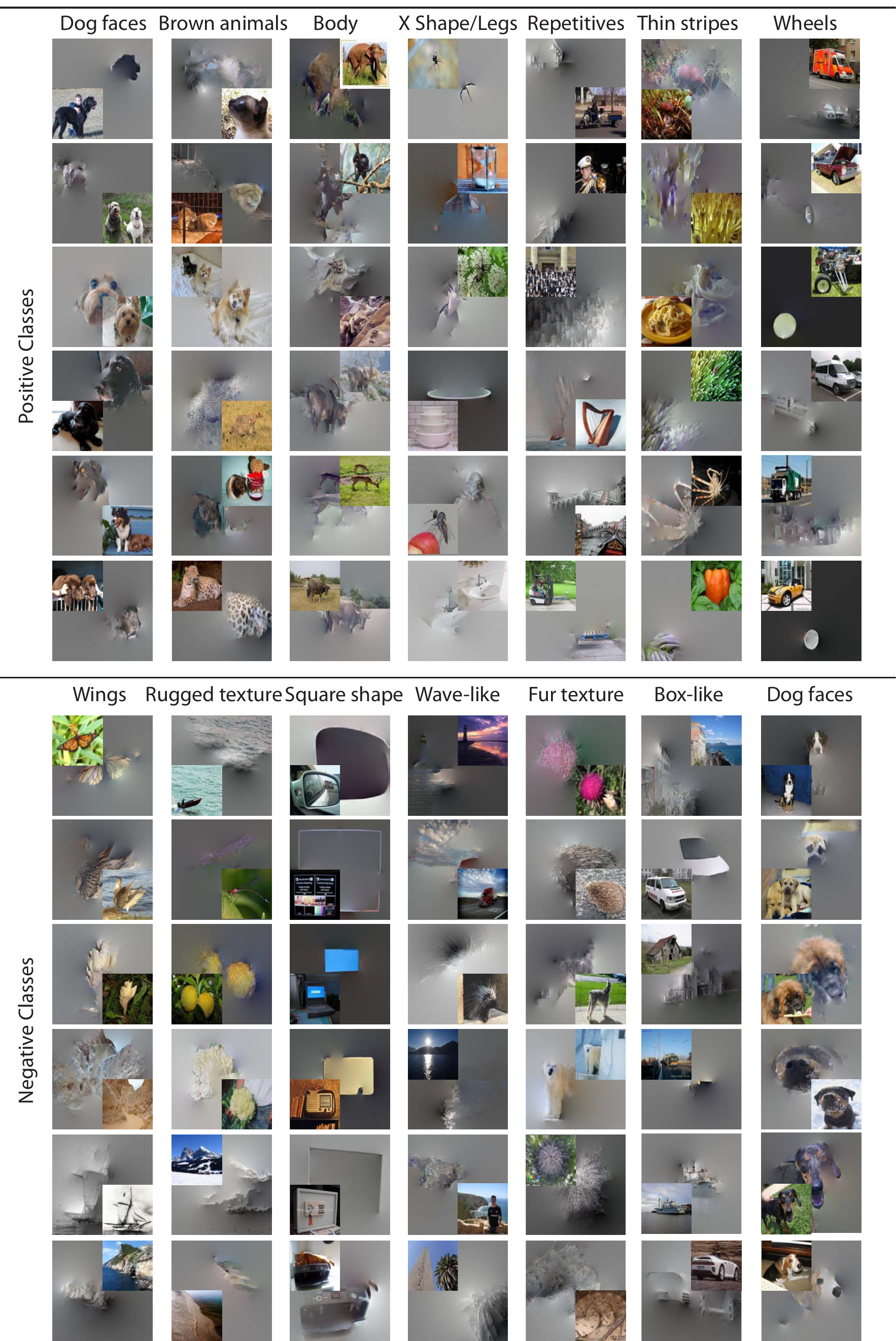}
\caption{Activation pattern for individual neurons. Each column represents a single neuron in the fc7 binary feature vectors. We show 6 example classes for positive and negative classes each. Each visualization is represented as the minimal image that causes the neuron to fire or inhibit. The original image is thumbnailed in the corner. We summarize the visual consistency across classes by our descriptions.}
\label{fig:neuron_meaning}
\vspace{-3mm}
\end{figure}

In figure \ref{fig:fire_pattern}, for neuron 1 and 2, the positive and negative classes are quite obviously corresponding to semantics, since ImageNet labels are based on WordNet hierarchy. But for other neurons, like neuron 3 and 4, this does not always seem to be the case. What visual properties do these classes share in order to fire/inhibit for the same neuron? For each class in the ``positive classes" and ``negative classes", we randomly select one image instance in the class, and find the minimum image representation to highlight the object part that contribute most to this neuron. We use the visualization technique from \cite{zhou2014object}. Concretely, we first segment the image by edges, regions and corners, and then remove each segment greedily until the binary response flips. In figure \ref{fig:neuron_meaning}, we show examples of 7 neurons. Please note that each image does not simply represent for an instance, but for the whole class. Unlike previous visualizations that attempt to find the consistency within a class, we show the consistent image pattern across classes. Our results show that some fc7 neurons do represent high level concept, like dogs, animals, and car wheels. While some others may still capture low level information like texture and shapes. One interesting result of pattern ``Wings" reveals the visual consistency of butterfly wings, goose wings, mushroom wings, sailboat wings, and even mountain wings. Moreover, \cite{parizi2014automatic} puts forward the notion of negative parts for classification. Our binary representation experimentally proves that negative parts automatically emerge in deep neural networks, and the same part can be shared as positive responses and negative responses such as dog faces in figure \ref{fig:neuron_meaning}.

%Maybe Todo: Dead Neuron, what leads to neurons that constantly fire or inhibit. Sparsity, histogram of number of neurons that corresponds to the same sparsity.

%\subsection{Quantifying compositionality of deep architectures}

\subsection{Towards binarizing model parameters}
We always lose some performance using binary representations under the same architecture. However, the number of model parameters remain the same. This means that there should be a lot more redundancy in the model with binary representations. Intuitively, changing a particular weight of the model does not necessarily change the output because the value is thresholded to binary. Also, the scale of the weights does not influence the output. Overall, parameters using a binary representation are less sensitive to the output responses. Can we compress the model parameters to binary altogether? Specifically, after we obtain the model with binary representations, we simply threshold the model weights to $-1, +1$, and keep it fixed. We leave the conv1 parameters to be real values since the input image is not binary and has rich sensory information. Then we finetune the bias and slope parameters using back-propagation. In this way, all the middle representations as well as most of the model parameters are binarized.

We conduct the experiments on CIFAR10 dataset. For the baseline architecture, we achieve 63.55\%, which is 11.85\% worse than the regular model 75.40\%, and 9.53\% worse than our binary representation model with real-valued parameters 73.08\%. For the twice larger network architecture, the performance increases to 69.68\%. To achieve the baseline accuracy, we make the network 4 times larger, and it finally reaches to 75.07\%. The model contains about 16 times numbers of model parameters and 4 times numbers of feature units compared with the baseline model. However, all the computations could be stored and calculated in a bit-wise way.
%The experiment shows that when a robust binary represetation is already attained, the model p

\vspace{-2mm}
\section{Conclusions}
\vspace{-3mm}
In this paper, we propose adjustable bounded rectifiers along with its optimization techniques to learn binary representations for deep neural networks.
Our results confirm the redundancy in current deep models. We can safely binarize the last layer and even all the layers with reasonable good performance. We also show our activation function can be used as a new way to regularize the model even when a binary representation is not desired. Using the learned representations, we can interpret the function of each binary neuron using semantic phrases by examining the minimal activation pattern of input images. We believe our binary model has great practical value for deployment, especially in targeted hardwares. In the future, we plan to optimize the smallest binary architecture that matches the state-of-the-art real value networks and incorporate binarizing model parameters into optimization.

%\subsubsection*{Acknowledgments}
%We thank Tong Xiao, Chen Huang, Aditya Khosla for inspiring discussions. We are grateful to Bolei Zhou and Agata Lapedriza for providing their visualization code, and Yuanjun Xiong for sharing his multi-GPU implementation. 
\clearpage
\bibliography{iclr2016_conference}
\bibliographystyle{iclr2016_conference}

\end{document}